\documentclass{segabs}

\date{} 

\usepackage{enumitem}

\usepackage{url}

\usepackage{hyperref} 

\begin{document}

\onecolumn 

\begin{description}[labelindent=0cm,leftmargin=3cm,style=multiline]

\item[\textbf{Citation}]{Z. Long, Y. Alaudah, M. Qureshi, M. Farraj, Z. Wang, A. Amin, M. Deriche, and G. AlRegib, ``Characterization of migrated seismic volumes using texture attributes: a comparative study,'' Proceedings of the SEG 85th Annual Meeting, New Orleans, LA, Oct. 2015.}

\item[\textbf{DOI}]{\url{https://doi.org/10.1190/segam2015-5934664.1}}

\item[\textbf{Review}]{Date of publication: October 2015}

\item[\textbf{Data and Codes}]{\url{https://ghassanalregibdotcom.files.wordpress.com/2016/10/zhiling_seg2015_code.zip}}

\item[\textbf{Bib}] {@incollection\{long2015characterization,\\
  title=\{Characterization of migrated seismic volumes using texture attributes: a comparative study\},\\
  author=\{Long, Z. and Alaudah, Y. and Qureshi, M. and Farraj, M. and Wang, Z. and Amin, A. and Deriche, M. and AlRegib, G.\},\\
  booktitle=\{SEG Technical Program Expanded Abstracts 2015\},\\
  pages=\{1744--1748\},\\
  year=\{2015\},\\
  publisher=\{Society of Exploration Geophysicists\}\}\\
} 


\item[\textbf{Copyright}]{\textcopyright 2015 SEG. Personal use of this material is permitted. Permission from SEG must be obtained for all other uses, in any current or future media, including reprinting/republishing this material for advertising or promotional purposes, creating new collective works, for resale or redistribution to servers or lists, or reuse of any copyrighted component of this work in other works.}

\item[\textbf{Contact}]{\href{mailto:zhiling.long@gatech.edu}{zhiling.long@gatech.edu}  OR \href{mailto:alregib@gatech.edu}{alregib@gatech.edu}\\ \url{https://ghassanalregib.com/} \\ }
\end{description}

\thispagestyle{empty}
\newpage
\clearpage
\setcounter{page}{1}

\twocolumn

\title{Characterization of migrated seismic volumes using texture attributes: a comparative study}

\renewcommand{\thefootnote}{\fnsymbol{footnote}}

\author{Zhiling Long\footnotemark[1], Yazeed Alaudah, Muhammad Ali Qureshi, Motaz Al Farraj, Zhen Wang, Asjad Amin, Mohamed Deriche, and Ghassan AlRegib\\
Center for Energy and Geo Processing (CeGP) at Georgia Tech and KFUPM}

\footer{Example}
\lefthead{Long et al.}

\maketitle

\begin{abstract}
In this paper, we examine several typical texture attributes developed in the image processing community in recent years with respect to their capability of characterizing a migrated seismic volume. These attributes are generated in either frequency or space domain, including steerable pyramid, curvelet, local binary pattern, and local radius index. The comparative study is performed within an image retrieval framework. We evaluate these attributes in terms of retrieval accuracy. It is our hope that this comparative study will help acquaint the seismic interpretation community with the many available powerful image texture analysis techniques, providing more alternative attributes for their seismic exploration. 
\end{abstract}

\section{Introduction}
In image processing, texture attributes are quantities generated from a texture pattern that capture the unique spatial distribution of the pixel intensities \cite[]{GW06}. Such texture attributes have been employed in some seismic interpretation applications. For instance, the gray level co-occurrence matrix (GLCM) was applied to salt dome detection \cite[]{berthelot2013texture} and deep-marine facies discrimination \cite[]{gao2007application}; Hilbert transform features were utilized for seismic image segmentation \cite[]{pitas1992texture}; Gabor filters were adopted for seismic image segmentation as well \cite[]{roster1998system}.

Although these applications were successful, we believe texture attributes can be even more useful for seismic exploration. Given that migrated seismic volumes are textural in nature, texture attributes have the potential of serving as local descriptors that characterize the migrated data. Such descriptors are essential for a computer-assisted understanding of the "subsurface scene," which can help pinpoint spots that are of more interest to an interpretor. In recent years, many powerful texture analysis techniques have been developed in the image processing community. Most of them have not been exposed adequately to exploration geophysicists yet. Therefore, in this paper, we conduct a comparative study examining several typical texture attributes with respect to their capability of characterizing a seismic volume. Please note that, although some of the techniques discussed in this paper may have been introduced to the community before, they will be explored in a different context, i.e., as texture attributes for general description of migrated seismic volumes.

The attributes we examine here can be categorized into frequency or space domain techniques. For the frequency domain, we explore the steerable pyramid (SP) \cite[]{119725}, and the curvelet transform (CT) \cite[]{candes05}. These are two typical extensions of the standard wavelet transform (WT), which is probably the most popular technique for spectral content-based image and texture analysis. SP improves over WT \cite[]{chui2014introduction} in that it achieves translation-invariance and orientation-invariance by dropping the orthogonality constraint. In addition to SP, several other multi-scale techniques have been developed more recently to overcome an inherent shortcoming of WT, which is the lack of directionality. Among such techniques, CT gained popularity particularly for seismic data processing, mainly because it describes with high efficiency signals of smooth curves.

For the space domain, we study the local binary pattern (LBP) \cite[]{ojala2002multiresolution} and the local radius index (LRI) \cite[]{Zhai2013:LRI}. LBP generates binary numbers in local areas according to gray-scale intensity differences among image pixels. The pattern of the binary numbers is captured for a local area as an occurrence histogram of the numbers. Since its introduction, LBP (and its variants) has been widely used for very successful texture analysis due to its robustness and computational efficiency. LRI is one of the most recently developed techniques following the concept of local patterns. It measures distances between local edges along all directions. The resulting histograms serve very well as a local texture discriptor. We note that although some of the older techniques such as GLCM have been used for seismic applications for a long time, they are not included in this paper. The major reason is that, first, they are already familiar to the community; and second, the newer techniques are more powerful.

To determine the capability of the attributes for characterizing migrated seismic data, we perform retrieval experiments, where a dataset is searched to identify images that are similar to a given query image in content, or in seismic structure when our dataset is concerned. The structural similarity between seismic images is measured on the basis of the attributes. Thus, the attributes' capability of characterizing the data (or the seismic structures contained therein) can be evaluated in terms of retrieval accuracy. Higher accuracies indicate that the associated attributes are better able to distinguish between different seismic structures. To achieve a more informed comparison, we choose SeiSIM \cite[]{Long2015seisim} as our benchmark. This method considers both the frequency domain statistics obtained using SP, and the space domain statistics calculated from discontinuity maps instead of original seismic images. One of very few techniques designed for evaluating similarity between seismic images, SeiSIM was demonstrated to be capable of capturing differences in geological structures with reliability. We hope this comparative study will help acquaint the seismic interpretation community with the many available powerful image texture analysis techniques, providing more alternative attributes for their seismic exploration.


\section{Methods}

\subsection{Frequency domain techniques}

\subsubsection{Steerable pyramid (SP)}

SP is a multiresolution image representation developed by \cite{119725}. As illustrated in Figure~\ref{fig:pyramid}, the technique first decomposes a given image into a highpass subband and a lowpass subband. Then it processes the lowpass subband, obtaining a series of bandpass subbands and another lowpass subband. The bandpass subbands reveal image details along various orientations. The newly obtained lowpass subband is subsampled and then further processed in a similar manner to yield orientational details at a coarser spatial scale. Such recursive decomposition eventually yields a pyramid of subbands, representing the original image along different orientations at different scales. Histograms of the coefficients from the decomposition can be established for each subband, which capture the statistical characteristics of the coefficients. The histograms are further examined for retrieval purpose, details of which will be discussed later in the experiments.

\renewcommand{\figdir}{Fig} 
\plot{pyramid}{width=0.48\textwidth}
{Illustration of a SP decomposition with $K$ scales and $N$ orientations at each scale. In this paper, we set $K = 4$ and $N = 8$.}

\subsubsection{Curvelet transform (CT)}

\begin{figure}
	\centering
	\includegraphics[width=0.5\linewidth]{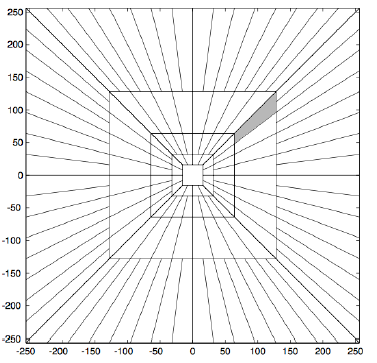}
	\caption{Curvelet tiling of the frequency spectrum showing different scales and orientations; adapted from \cite{candes05}.}
	\label{fig:curvelet}
\end{figure}

CT is also a directional multiscale decomposition, first introduced by \cite{candes05}. It works by first applying the two-dimensional fast Fourier transform (2-D FFT) to an image, and then dividing the frequency plane into small sections (or wedges) corresponding to multiple scales and orientations. The total number of scales in the curvelet tiling, $J$, is dependent on the size of the image as
\begin{equation}\label{nScales}
J =  \lceil \log_2 \min (N_1, N_2) -3 \rceil ,
\end{equation}
where $N_1$ and $N_2$ are the image height and width in pixels, respectively; and $\lceil \cdot \rceil$ is the ceiling function. The number of orientations at scale $j\geq 1$, $K(j)$, is given by:
\begin{equation}
K(j) = 16 \times 2^{\lceil (j-1)/2 \rceil}.
\end{equation}
Once the frequency plane is partitioned (see Figure \ref{fig:curvelet} for an example), curvelet coefficients are generated by applying the 2-D IFFT to each wedge (after smoothing). Since the FFT of real images is symmetric around the origin, only two quadrants of the Fourier spectrum are necessary for obtaining the coefficients. Again, histograms are formed for coefficients in each subband and used in the retrieval experiments.



\subsection{Space domain techniques}

\subsubsection{Local binary pattern (LBP)}

LBP \cite[]{ojala2002multiresolution} describes the local spatial structure of textures by thresholding the neighborhood of each pixel and defining the result as a binary number. Mathematically, the LBP operator is expressed as
\begin{equation}
\label{equ:lbp}
LBP_{R, P}\left[i_c,j_c\right]=\sum\limits_{p=0}^{P-1}s\left(I_c-I_p\right)\cdot 2^p,
\end{equation}
where $P$ represents the number of points in the neighborhood with radius $R$, $\left[i_c,j_c\right]$ indicates the coordinates of the center point, and $I_c$ and $I_p$ denote the intensity of the center and neighboring points, respectively. Function $s(\cdot)$ has a value of $1$ if $I_c\geq I_p$. Otherwise, the value of $s$ is $0$. Since the LBP operator encodes only the signs of the difference between the center and neighboring points, however, the information of difference magnitude has been discarded. To overcome this problem, \cite{guo2010completed} proposed completed LBP (CLBP), where three components are considered as follows. First, CLBP\_C encodes the center pixel intensity into a binary number. Then, CLBP\_S and CLBP\_M are generated using the difference between the center and its neighbors, with the former encodes the sign of the difference and the latter the magnitude. Histograms of the three components are concatenated into one feature vector to describe the local texture pattern. In fact, CLBP\_S is exactly the same as LBP. In this paper, we use CLBP instead of the original LBP. We set $P = 20$ and $R = 3$.

\subsubsection{Local radius index (LRI)}

\begin{figure}
   \begin{center}
   \begin{tabular}{cc}
   \includegraphics[width=0.45\linewidth]{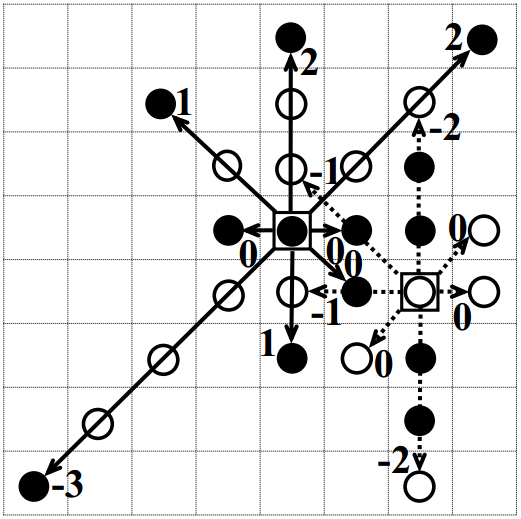} &
   \includegraphics[width=0.45\linewidth]{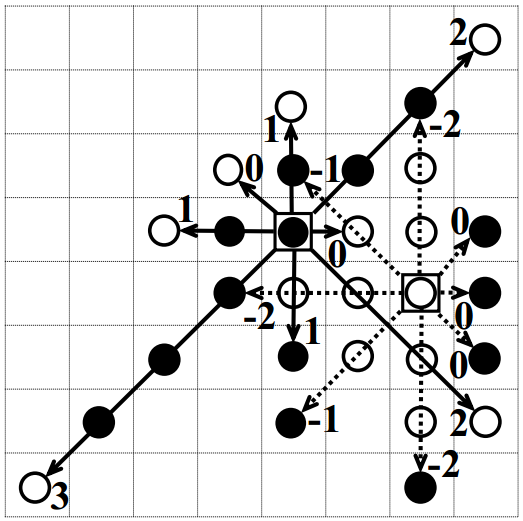} \\
   (a) & (b) \\
   \end{tabular}
   \end{center}
   \caption[LRI]
   { \label{fig:LRI}
    An example illustrating computation of (a) LRI-A and (b) LRI-D. The local indices are calculated at the two example points (one black and one white, within a square box) along 8 directions. A black point is an edge point.}
   \end{figure}

LRI characterizes a texture pattern by the distribution of distances between adjacent edges along a certain orientation \cite[]{Zhai2013:LRI}. A local index can be computed for each image pixel in two different ways, resulting in two variations of LRI. For LRI-A, inter-edge distance (i.e., width of adjacent smooth regions) in each given direction is calculated; while for LRI-D, the distance from pixels to the nearest edge (i.e., boundary of next smooth region) is adopted. In Figure \ref{fig:LRI}, an example is given illustrating how to compute LRI-A and LRI-D. Histograms calculated from LRI-A and LRI-D are concatenated to form one vector for the retrieval experiments.

There are two key parameters for LRI calculation. One is a threshold $T$ to determine edge. The other is a range $K$ to control how far to search for edges. In our experiments, we use $T = \sigma/2$ (where $\sigma$ is the local standard deviation) and $K = 3$. We obtain a $(2K+1)$-bin histogram in each of eight directions.

\section{Experiments}
\subsection{Data}

\begin{figure}
   \begin{center}
   \begin{tabular}{cc}
   \includegraphics[width=0.4\linewidth]{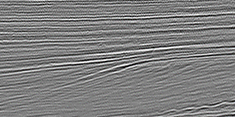} &
   \includegraphics[width=0.4\linewidth]{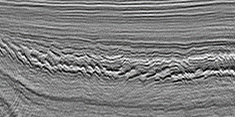} \\
   (a) & (b) \\
   \includegraphics[width=0.4\linewidth]{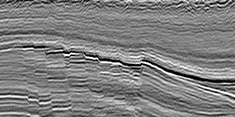} &
   \includegraphics[width=0.4\linewidth]{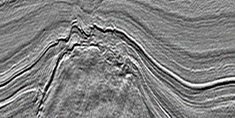} \\
   (c) & (d) \\
   \end{tabular}
   \end{center}
   \caption[exampleImages]
   { \label{fig:exampleImages}
    Example images illustrating different geological structures in the dataset: (a) clear horizon, (b) chaotic horizon, (c) faults, and (d) salt dome.}
   \end{figure}

We created a dataset consisting of 400 images extracted from the public dataset of Netherlands offshore F3 block with the size of $24\times16 km^2$ in the North Sea. The images are grouped into 4 classes according to their geological structures, namely, clear horizon, chaotic horizon, faults, and salt dome. Each group includes 100 images. All images are $150 \times 300$ in pixels. Example images are given in Figure \ref{fig:exampleImages}.

\subsection{Experimental procedure}

We conduct retrieval experiments on the dataset where the similarity is measured based on the various texture attributes discussed above. For a complete evaluation, every image in the dataset is selected once as the query image. It is then compared with the rest of images, ranking them according to the similarity score in a descending order. Such ranking is obtained for all images. Then the performance can be assessed using all rankings.

The images are first normalized. Then for each image, a specific texture attribute is computed and histograms are formed for the attributes. The images are compared in terms of a certain distance measure calculated between corresponding histograms. We used squared chord distance (SCD) \cite[]{similarity_measures} to generate all the results presented in this paper. We also tested other distance measures such as the Kullback-Leibler divergence (KLD) \cite[]{Kullback1951:KLD}, and noticed no significant difference in the results. SCD is defined as:
\begin{equation}
d_{SC}(H^f, H^g) = \sum_{i = 1}^M \left( \sqrt{H^f(i)} - \sqrt{H^g(i)} \right) ^2 ,
\end{equation}
where $H^f$ and $H^g$ are the two histograms to be compared, from images or subbands $f$ and $g$, respectively. $H^f(i)$ (or $H^g(i)$) is the $i^{th}$ bin in the histogram $H^f$ (or $H^g$). $M$ is the number of bins in the histogram.

The overall distance between the two images is given by combining SCD calculated over all $J$ pairs of histograms:
\begin{equation}
D(f,g) = \sum_{j=1}^{J-1} d_{SC}(H^f_{j}, H^g_{j})
\end{equation}
The overall distance can then be easily converted to similarity values bounded between $0$ and $1$.

\subsection{Evaluation metrics}

We adopt the following metrics that are typically used to evaluate retrieval performance (\cite{STSIM,Zhai2013:LRI}). They are all in the range of $0-1$, with the higher value indicating better performance.

\begin{itemize}
\item \textbf{Precision at $n$ (P@$n$)}

This index presents the precision up to the $n^{th}$ retrieved image.

\item \textbf{Mean average precision (MAP)}

This index accounts for the case when multiple matching images are in existence in the dataset. For each query image, the database is retrieved till the last matching image is identified. For a certain matching image of which the rank is $n$, the fraction of total matching images among the first $n$ retrieved images is used as the associated precision. The average retrieval precision for a query is calculated using the precision associated with each of its matching images. Then, the MAP is obtained by averaging across all query images.

\item \textbf{Retrieval accuracy (RA)}

This is the overall retrieval accuracy.

\item \textbf{Area under curve (AUC)}

This index calculates the area under the curve of the receiver operating characteristic (ROC), which is typical for binary detection problem considering both the rate of true detection and the rate of false positive at the same time.

\end{itemize}

\subsection{The benchmark: SeiSIM}

SeiSIM is a metric that was proposed to evaluate similarity between two migrated seismic images \cite[]{Long2015seisim}. It combines frequency domain texture attributes with space domain geological attributes. In the frequency domain, texture similarity is evaluated using statistics calculated on subbands decomposed from SP, an approach denoted as STSIM-1 \cite[]{STSIM}.
\begin{equation}
Q_{STSIM-1}(\textbf{x},\textbf{y})=[l(\textbf{x},\textbf{y})]^\frac{1}{4}[c(\textbf{x},\textbf{y})]^\frac{1}{4}[a_h(\textbf{x},\textbf{y})]^\frac{1}{4}[a_v(\textbf{x},\textbf{y})]^\frac{1}{4}
\end{equation}
where $l(\textbf{x},\textbf{y})$ is the term accounting for luminance similarity, $c(\textbf{x},\textbf{y})$ represents contrast similarity, $a_h(\textbf{x},\textbf{y})$ is the structural similarity based on horizontal autocorrelation, and $a_v(\textbf{x},\textbf{y})$ gives the structural similarity according to the vertical autocorrelation, all calculated between subband images $\textbf{x}$ and $\textbf{y}$.

In the space domain, geological similarity, denoted as DM \cite[]{Long2015seisim}, is assessed using statistics obtained from discontinuity maps associated with the original images. The similarity between two discontinuity maps $DM_1$ and $DM_2$ is then defined as
\begin{equation}
Q_{DM}(DM_1,DM_2)=[a_{DM}^h(DM_1,DM_2)]^\frac{1}{2}[a_{DM}^v(DM_1,DM_2)]^\frac{1}{2}
\end{equation}
where $a_{DM}^h(DM_1,DM_2)$ finds the similarity based on horizontal autocorrelation calculated in each map, and $a_{DM}^v(DM_1,DM_2)$ determines the similarity using vertical autocorrelation from each map.

Consequently, the seismic similarity is expressed as
\begin{equation}
SeiSIM=[Q_{STSIM-1}]^\frac{1}{2}[Q_{DM}]^\frac{1}{2}.
\end{equation}
When applied to the retrieval problem in this paper, SeiSIM does not follow the framework used for the other texture attributes discussed earlier. It extracts simple statistics such as means, variances, and autocorrelations from the attributes and directly compares them between images. In contrast, the retrieval framework in this paper for all other techniques forms histograms of the attributes and compares them instead. Obviously, SeiSIM is not exactly comparable to the others. Therefore, we use it in this study as a benchmark, since it has been demonstrated to be a good measure of the seismic similarity \cite[]{Long2015seisim}.

\subsection{Results}

\begin{table}
\centering
\caption{Retrieval performance for different attributes.}
\label{table:results}
\begin{tabular}{lccccccc}
\hline
 & P@20 & P@50 & MAP & RA & AUC\\
\hline
\hline
SP & 1.000 & 0.965 & 0.928 & 0.866 & 0.965\\
CT & 1.000 & 0.968 & 0.954 & 0.914 & 0.988\\
LBP & 0.999 & 0.953 & 0.932 & 0.871 & 0.967\\
LRI & 0.997 & 0.977 & 0.953 & 0.896 & 0.968\\
SeiSIM & 1.000 & 0.992 & 0.974 & 0.943 & 0.990\\
\hline
\end{tabular}
\end{table}

The retrieval results are shown in Table \ref{table:results}. The overall best performance is observed with SeiSIM, for which we believe the incorporation of the spatial geological attributes (discontinuity) plays an important role. As we pointed out earlier, SeiSIM is not exactly a comparable method within the framework adopted in this paper, but used as a benchmark. Comparing with SeiSIM, generally the retrieval performance was excellent for the four attributes being examined. P@20 values are all perfect. P@50 values are close to each other among the four, and still close to SeiSIM. The same observation is made with AUC. For MAP, CT and LRI yield higher values than SP and LBP. While for RA, CT is the best, followed by LRI. RA results for SP and LBP are very close and much lower than CT. Among the four attributes, CT yields the best overall performance, closest to SeiSIM. LRI is the second best when all metrics are considered. We conclude that the attributes all demonstrate great potential of effectively capturing the characteristics of the different geological structures, with CT and LRI being the most promising ones.

We also examined the computational time required for comparing a pair of images using the techniques. The comparison was performed on a computer with the following configuration: Intel Core i7, 3.4 GHz, with 32GB of RAM, and running on 64-bit windows 7. The results are shown in Table \ref{table:results2}. CT is the fastest among all techniques.

\begin{table}
\centering
\caption{Comparison of computation time in seconds.}
\label{table:results2}
\begin{tabular}{ccccc}
\hline
SP & LBP & LRI & CT & SeiSIM\\
\hline
\hline
0.2137 & 0.1638 & 2.3946 & 0.1201 & 1.1872\\
\hline
\end{tabular}
\end{table}

\section{Conclusions}
In this paper, we conducted a comparative study of several typical examples of texture attributes developed in the image processing community in recent years, covering different techniques in frequency and space domain. Within a framework for image retrieval, the attributes were examined in terms of retrieval accuracy. The study demonstrated that texture attributes are generally capable of characterizing a migrated seismic volume according to its geological structure, thus can be effective for computer-assisted understanding of subsurface structures. We hope this study will also introduce to the exploration geophysicists the existing powerful image texture analysis methods, which have the potential to provide useful attributes for various seismic exploration applications.

\section{ACKNOWLEDGMENTS}
This work is supported by the Center for Energy and Geo Processing (CeGP) at Georgia Tech and King Fahd University of Petroleum and Minerals (KFUPM). 

\bibliographystyle{seg}  
\bibliography{main}

\end{document}